\documentclass[article]{jss}\usepackage[]{graphicx}\usepackage[]{xcolor}
\makeatletter
\def\maxwidth{ %
  \ifdim\Gin@nat@width>\linewidth
    \linewidth
  \else
    \Gin@nat@width
  \fi
}
\makeatother

\usepackage{Sweave}

\usepackage{thumbpdf}
\usepackage{lmodern}
\usepackage{amsmath}
\usepackage{amssymb}
\usepackage{tabls}

\newcommand{\indep}{\hspace{0.1em}\perp\hspace{-0.95em}\perp}
\newcommand{\given}{\mid}

\newcommand{\y}{\mathbf{y}}
\newcommand{\wy}{\widehat{\y}}
\newcommand{\Xm}{\mathbf{X}}
\newcommand{\Sm}{\mathbf{S}}
\newcommand{\si}{\mathbf{s}_i}
\newcommand{\sj}{\mathbf{s}_j}
\newcommand{\U}{\mathbf{U}}
\newcommand{\wU}{\widehat{\U}}
\newcommand{\balpha}{\boldsymbol{\alpha}}
\newcommand{\bbeta}{\boldsymbol{\beta}}
\newcommand{\be}{\boldsymbol{\varepsilon}}
\newcommand{\B}{\mathbf{B}}
\newcommand{\T}{\mathrm{T}}
\newcommand{\argmin}{\operatornamewithlimits{argmin}}
\newcommand{\OLS}[1]{\widehat{#1}_{\mathrm{OLS}}}
\newcommand{\NCLM}[1]{\widehat{#1}_{\mathrm{NCLM}}}
\newcommand{\FRRM}[1]{\widehat{#1}_{\mathrm{FRRM}}}
\newcommand{\FGRRM}[1]{\widehat{#1}_{\mathrm{FGRRM}}}
\newcommand{\lr}{\lambda(r)}
\newcommand{\rs}{R^2_\Sm}
\newcommand{\FaVa}{\FRRM{\balpha}^\T \VAR(\Sm) \FRRM{\balpha}}
\newcommand{\FbVb}{\FRRM{\bbeta}^\T \VAR(\wU) \FRRM{\bbeta}}
\newcommand{\st}{\;\;\text{such that}\;\;}
\newcommand{\mref}[1]{(\ref{#1})}
\newcommand{\Zm}{\mathbf{0}}
\newcommand{\I}{\mathbf{I}}

\author{Marco Scutari \\
 Istituto Dalle Molle di Studi sull'Intelligenza Artificiale (IDSIA)}
\Plainauthor{Marco Scutari}

\title{\pkg{fairml}: A Statistician's Take on Fair \\ Machine Learning Modelling}
\Plaintitle{fairml: A Statistician's Take on Fair Machine Learning Modelling}
\Shorttitle{fairml: A Statistician's Take on Fair Machine Learning Modelling}

\Abstract{
  The adoption of machine learning in applications where it is crucial to ensure
  fairness and accountability has led to a large number of model proposals in
  the literature, largely formulated as optimisation problems with constraints
  reducing or eliminating the effect of sensitive attributes on the response.
  While this approach is very flexible from a theoretical perspective, the
  resulting models are somewhat black-box in nature: very little can be said
  about their statistical properties, what are the best practices in their
  applied use, and how they can be extended to problems other than those they
  were originally designed for. Furthermore, the estimation of each model
  requires a bespoke implementation involving an appropriate solver which is
  less than desirable from a software engineering perspective.

  In this paper, we describe the \pkg{fairml} R package which implements our
  previous work \citep{stco21} and related models in the literature.
  \pkg{fairml} is designed around classical statistical models (generalised
  linear models) and penalised regression results (ridge regression) to produce
  fair models that are interpretable and whose properties are well-known. The
  constraint used to enforce fairness is orthogonal to model estimation, making
  it possible to mix-and-match the desired model family and fairness definition
  for each application. Furthermore, \pkg{fairml} provides facilities for model
  estimation, model selection and validation including diagnostic plots.

}

\Plainkeywords{fairness, ridge regression, generalised linear models}

\Address{
  Marco Scutari \\
  Istituto Dalle Molle di Studi sull'Intelligenza Artificiale (IDSIA) \\
  Polo Universitario Lugano, Campus Est \\
  Via la Santa 1 \\
  6962 Lugano, Switzerland \\
  E-mail: \email{scutari@bnlearn.com} \\
  URL: \url{https://www.bnlearn.com/}
}
\IfFileExists{upquote.sty}{\usepackage{upquote}}{}
\begin{document}

\section{Introduction}
\label{sec:intro}

In many applications it is crucial to ensure the interpretability,
explainability and fairness of the decisions made on the basis of machine
learning models: some examples are criminal justice \citep{heidari}, credit risk
modelling \citep{credit} and screening job applications \citep{jobs}. Failure to
do so has resulted in discrimination based on race \citep{compas}, gender
\citep{stem-ads}, nationality \citep{castillo} and age \citep{diaz,hort} among
others. As a result, the US, the EU and the UK have recently introduced
legislation to regulate the use of machine learning models \citep{cath}. The
European Commission has also released the first legal framework for the use of
artificial intelligence \citep{eu2021}, which is now under revision by the
member states. On a broader scale (both geographical and temporal), improving
fairness and reducing inequality are integral to the United Nations' Sustainable
Development Goals \citep{sdgs}.

The introduction of legal requirements and initiatives like AI for Good
\citep{aiforgood} have spurred the development of \emph{algorithmic fairness} as
an independent research field. In addition to different mathematical
characterisation of fairness, many models and algorithms have been proposed by
leveraging (constrained) optimisation and information theory to achieve the best
possible predictive accuracy while ensuring that we are not discriminating
individuals based on sensitive (or legally restricted) attributes. These models
and algorithms are typically limited to an objective functions and a set of
constraints, representing goodness of fit and fairness respectively, instead of
being full-fledged probabilistic models: their statistical properties, best
practices for model selection and validation, significance testing etc. would
have to be re-derived from scratch for each of them. In our previous work
\citep{stco21}, we have taken the opposite view that classical statistical
models can be adapted to enforce fairness while preserving their well-known
properties and the associated best practices in their applied use. We showed
that combining generalised linear models (GLMs) with penalised regression works
very well for this purpose, and we are now providing a production-grade
implementation of these and related models for \proglang{R} \citep{rcore} in the
\pkg{fairml} package \citep{fairml}. \proglang{R} provides a rich environment
for statistical modelling to integrate into, and a well-structured and tested
\proglang{R} package is more reliable and suitable for general use than the
\proglang{Python} proof-of-concept scripts typically available in the
algorithmic fairness literature.

The aim of this paper is to provide an overview of \pkg{fairml}. In
Section~\ref{sec:models}, we will briefly review the methodological literature
on fair machine learning models (Section~\ref{sec:fair-reg}) and the
availability of software packages (Section~\ref{sec:software}), focusing in
particular on linear models. In Section~\ref{sec:fgrrm}, we will introduce
the fair ridge regression (FRRM) and fair generalised ridge regression (FGRRM)
models from our previous work \citep{stco21} along with the other models
implemented in \pkg{fairml}. In Section~\ref{sec:features}, we will then
describe the software architecture and the features of the package. Finally, we
will illustrate the use of relevant functions in Section~\ref{sec:showcase} and
summarise our conclusions in Section~\ref{sec:discussion}.

\section{Fair Machine Learning Models and Software}
\label{sec:models}

Algorithmic fairness research comprises two complementary topics: exploring
mathematical characterisations of fairness and efficiently estimating models
that produce fair predictions.

The variety of fairness criteria and characterisations in the literature have
been recently reviewed in \citet{mehrabi}, \citet{delbarrio} and
\citet{pessach}. Broadly speaking, they follow two approaches: \emph{group} and
\emph{individual fairness}. Group fairness criteria require predictions to be
similar across the groups identified by the sensitive attributes. They are
typically expressed, for a given model,  as conditional independence statements
of the fitted values $\wy$ for some response variable $\y$ from the sensitive
attributes $\Sm$. Two popular examples are \emph{statistical} or
\emph{demographic parity} ($\wy \indep \Sm$) and \emph{equality of opportunity}
($\wy \indep \Sm \given \y$).
Individual fairness criteria, on the other hand, require that individuals that
are similar receive similar predictions and are expressed as the cumulated
difference in the predictions ($d_1(\cdot)$ below) between pairs individuals
$(i, j)$ in different groups (identified by $d_2(\cdot)$ below):
\begin{equation}
  f(\balpha \y, \Sm) =
    \sum\nolimits_{i, j} d_1(y_i, y_j) d_2(\si\balpha, \sj\balpha).
\label{eq:individual}
\end{equation}
Both group fairness and individual fairness have originally been defined in the
simple scenario in which $\Sm$ contains a single, binary sensitive attribute,
but have been extended to multiple and to continuous sensitive attributes for
specific models \citep[see, for instance,][]{komiyama,zafar}.

As for fair modelling, there have been numerous attempts to address
discrimination at different stages of the model selection, estimation and
validation process, and for different classes of models. For the former, we can
distinguish \citep{lagatta}:
\begin{itemize}
  \item \emph{Pre-processing} approaches that try to transform the data to
    remove the underlying discrimination so that any model fitted on the
    transformed data is guaranteed to be fair. A foundational work that
    takes this approach is \citet{calmon}, which describes how to learn a
    probabilistic transformation that alters data towards group and individual
    fairness while penalising large feature changes in order to preserve data
    fidelity.
  \item \emph{In-processing} approaches that modify the model estimation process
    in order to remove discrimination, either by changing its objective function
    (typically the log-likelihood) or by imposing constraints on its parameters.
  \item \emph{Post-processing} approaches that use a hold-out set to assess a
    previously-estimated model (treated as a black box) and that alter its
    predictions to make them fair. \citet{moritz}, for instance, show how fair
    Bayes-optimal predictors can be derived from non-fair predictors from
    (binary) classifiers.
\end{itemize}

The nature of in-processing approaches depends strongly on the type of model we
are estimating. Learning fair black-box machine learning models such as deep
neural networks is most challenging \citep[see, for instance, ][]{choras} and is
being investigated in the broader Explainable AI community \citep{xai}. For this
reason, a large part of the literature focuses on simpler models. In many
settings, such models are preferable because there are limited amounts of data,
because of computational limitations or because they are more interpretable.
This the case for the \pkg{fairml} package and for our previous work
\citep{stco21}: we will provide a more detailed overview of these types of fair
models in the next section.

\subsection{Related Work on Fair Regression and Classification Models}
\label{sec:fair-reg}

All classical linear models in common use have been adapted in the literature to
enforce fairness: fair classification models are more common than fair
regression models, which are in turn more common than other families of GLMs.

Classification models are typically based on (binary) logistic regression: for
instance, \citet{woodworth} directly constrain equality of opportunity for a
binary sensitive attribute; and \cite{zafar} investigate the unfairness of the
decision boundary in logistic regression and support-vector machines under
statistical parity. \citet{reductions} use ensembles of logistic regressions
and gradient-boosted trees to reduce fair classification to a sequence of
cost-sensitive classification problems. Multinomial logistic regression has not
been also investigated, but not to the same extent because its formulation is
more complicated \citep{narasimhan, cotter}.

As for fair linear regression models, fairness has been enforced using auxiliary
generative models \citep{fukuchi}, clustering \citep{calders} and kernel
regularisation \citep{suay}. These and other approaches \citep{grouploss,chzhen}
leverage the literature on fair classification by discretising either the
response variable, the sensitive attributes or both; or by limiting models to a
single discrete (often binary) sensitive attributes. Likewise, many of these
methods define fairness as equality of opportunity; \citet{berk} is a notable
exception that enforces both individual and group fairness.

Two fair regression works (other than ours) are implemented in \pkg{fairml:}
\citet{komiyama} and \citet{zafar}. Both define fairness as statistical parity.
The former proposes a quadratic optimisation approach for fair linear regression
models that constrains least squares estimation by bounding the relative
proportion of the variance explained by the sensitive attributes. The latter
bounds the correlation between individual sensitive attributes and the fitted
values. These two approaches have several advantages over those mentioned above:
\begin{itemize}
  \item both predictors and sensitive attributes are allowed to be continuous as
    well as discrete (encoded with contrasts);
  \item any number of predictors and sensitive attributes can be included in the
    model;
  \item the level of fairness can be controlled directly by the user, without
    the need of model calibration to estimate it empirically;
  \item fairness is mathematically codified in the same way as the model's loss
    function, which means that there is no risk that the effects of sensitive
    attributes will accidentally be captured by model estimation.
\end{itemize}

\citet{do} and \cite{stco21} take a similar approach and extend it to span GLMs
to encompass both linear regression and classification: the former using a
first-order approximation of the GLM log-likelihood, the latter with a ridge
penalty (as we will discuss more in depth in Section \ref{sec:fgrrm}).

\subsection{Related Software}
\label{sec:software}

At the time of this writing, there isn't much in the way of software packages to
work with fair models. Outside of the \proglang{R} ecosystem, IBM AI Fairness
360 \citep{ai360} is the only choice that provides a wide set of pre-processing
and post-processing methods, along with some in-processing methods.

As for \proglang{R}, the \pkg{mlr3fairness} \citep{mlr3fairness} package extends
the \pkg{mlr3} R package \citep{mlr3} with pre-processing and post-processing
methods, and provides a front-end to \pkg{fairml} for in-processing methods. The
\pkg{fairness} \citep{fairness} and \pkg{fairmodels} \citep{fairmodels}
packages implement computation and visualisation of fairness metrics.
\pkg{predfairness} \citep{predfairness} post-processes model predictions.

\section{Fair Generalised Ridge Regression Models}
\label{sec:fgrrm}

The core of \pkg{fairml} are the \emph{fair ridge regression model} (FRRM) and
the \emph{fair generalised ridge regression models} (FGRRM) from \citet{stco21}.
FRRM is defined as
\begin{align}
  &\y = \mu + \Sm\balpha + \wU\bbeta + \be&
  &\text{with}&
  &\wU = \Xm - \B^\T \Sm,
\label{eq:frrm}
\end{align}
where $\y$ is the response variable, $\Xm$ is the design matrix of predictors,
$\Sm$ is the design matrix of sensitive attributes and $\be$ are the residuals
of the model. The matrix $\wU$ represents the components of the predictors $\Xm$
that are orthogonal to the sensitive attributes $\Sm$, computed by regressing
the former on the latter with ordinary least squares and taking the residuals.

The regression coefficients $\balpha$ associated with $\Sm$ are subject to a
ridge penalty to reduce the effect of the sensitive attributes on $\y$, while
the $\bbeta$ are not penalised because $\wU$ is orthogonal to $\Sm$ and
therefore free from discriminatory effects.\footnote{$\mu$ is not penalised
either, in keeping with standard practices in penalised regression models.} In
other words, $\balpha$ and $\bbeta$ are estimated as
\begin{equation}
  (\FRRM{\balpha}, \FRRM{\bbeta}) = \argmin_{\balpha, \bbeta}
    \| \y - \mu - \Sm\balpha - \wU\bbeta \|_2^2 + \lr \|\balpha\|_2^2
\label{eq:frrm2}
\end{equation}
where $\lr \in \mathbb{R}^+$ is the penalty coefficient that results in
\begin{equation}
  \rs(\FRRM{\balpha}, \FRRM{\bbeta}) =
    \frac{\VAR(\Sm\FRRM{\balpha})}{\VAR(\wy)} = \\
    \frac{\FaVa}{\FaVa + \FbVb}
\label{eq:nclm-bound}
\end{equation}
begin equal to a user-specified unfairness level $r \in [0,1]$. That is, the
user chooses how fair the model should be and FRRM internally selects the value
of $\lr$ that gives the best goodness of fit under this constraint. For a given
$\lr$, $\FRRM{\balpha}$ and $\FRRM{\bbeta}$ have the usual closed-form
expressions
\begin{align}
  &\FRRM{\balpha} = \left(\Sm^\T \Sm + \lr \I\right)^{-1} \Sm^\T \y,&
  &\FRRM{\bbeta} = (\wU^\T \wU)^{-1} \wU^\T \y
\label{eq:estimates}
\end{align}
which makes numerically finding $\lr$ from $r$ computationally efficient.

The level of unfairness is defined as the proportion of the variance of the
fitted values $\wy$ that is explained by the sensitive attributes $\Sm$: $r = 0$
($\lr \to +\infty$) corresponds to a completely fair model in which $\wy$ is
independent from $\Sm$, while $r = 1$ ($\lr = 0$) is an inactive constraint by
construction since $\rs \leqslant 1$. In practical applications, imposing $r =
0$ is typically unfeasible and low values of $r$ are preferred because they
provide a better trade-off between fairness and predictive accuracy.

The FRRM model in \mref{eq:frrm2} can be rewritten as an optimisation problem as
\begin{equation*}
  \min_{\balpha, \bbeta} \E\left[(\y - \wy)^2\right]
  \st \|\balpha\|_2^2 \leqslant t(r),
\end{equation*}
which is similar to the non-convex linear model (NCLM) proposed by
\citet{komiyama}:
\begin{equation}
  \min_{\balpha, \bbeta} \E\left[(\y - \wy)^2\right]
  \st \rs(\balpha, \bbeta) \leqslant r.
\label{eq:nclm}
\end{equation}
The latter directly constrains $\rs$ and estimates $\balpha$ and $\bbeta$ using
a quadratic-constraints quadratic programming solver instead of just
constraining $\balpha$. As a result, it cannot be easily extended into a GLM;
and the behaviour of the estimated $\NCLM{\bbeta}$ is non-intuitive because they
vary with $r$ even though $\wU$ is free from discriminatory effects.

On the other hand, it is straightforward to extend FRRM into FGRRM using the
literature on generalised ridge regression models \citep{glmenet,coxenet} and
replacing the proportion of variance explained by $\Sm$ with the corresponding
proportion of the deviance. That is, we estimate $\balpha$ and $\bbeta$ as
\begin{equation}
  (\FRRM{\balpha}, \FRRM{\bbeta}) = \argmin_{\balpha, \bbeta}
    D(\balpha, \bbeta) + \lambda(r) \|\balpha\|_2^2.
\label{eq:fgrrm}
\end{equation}
where $D(\cdot)$ is the deviance, and we choose $\lr$ such that
\begin{equation}
  \frac{D(\balpha, \bbeta) - D(\Zm, \bbeta)}
       {D(\balpha, \bbeta) - D(\Zm, \Zm)} \leqslant r.
\label{eq:glmbound}
\end{equation}
where $\Zm$ is a vector of zeroes. In particular:
\begin{itemize}
  \item For a Gaussian GLM, \mref{eq:fgrrm} is identical to \mref{eq:frrm2}
    because the deviance is just the residual sum of squares and
    \mref{eq:glmbound} simplifies to $\rs(\balpha, \bbeta) \leqslant r$. FGRRM
    simply reverts to FRRM.
  \item For a Binomial or Multinomial GLM with the canonical logistic link
    function, that is, a (multinomial) logistic regression, \mref{eq:glmbound}
    bounds the difference made by $\Sm\balpha$ in the classification odds.
  \item For a Poisson GLM with the canonical logarithmic link, that is, a
    log-linear regression, \mref{eq:glmbound} bounds the difference in the
    intensity (that is, the expected number of arrivals per unit of time).
  \item For Cox's proportional hazards model, we can write the hazard function
    as
    \begin{equation*}
      h(t; \wU, \Sm) = h_0(t)\exp(\Sm\balpha + \wU\bbeta)
    \end{equation*}
    where $h_0(t)$ is the baseline hazard at time $t$. The corresponding
    deviance can be used as in \mref{eq:fgrrm} and \mref{eq:glmbound} to enforce
    the desired level of fairness, bounding the ratio of hazards through the
    difference in the effects of the sensitive attributes.
\end{itemize}

Furthermore, F(G)RRM is completely modular in that it separates model estimation
and fairness enforcement. The parameter estimates $\{\FRRM{\balpha}$,
$\FRRM{\bbeta}\}$ ($\{\FGRRM{\balpha}$, $\FGRRM{\bbeta}\}$, respectively) only
depend on the constraint $\rs \leqslant r$ through $\lr$. As a result, any
fairness constraint which induces a monotonic relationship between $r$ and
$\lr$ can be plugged into F(G)RRM without any change to \mref{eq:estimates}.
The individual fairness constraint from \citet{berk} satisfies this condition,
and it is trivial to alter \mref{eq:nclm-bound} to create an
equality-of-opportunity constraint that does as well
(Section~\ref{sec:features}). This is unlike other methods in the literature,
which integrate the fairness constraint deeply into the estimation process
making it impossible to mix-and-match model families and fairness constraints.

NCLM, FRRM and FGRRM constrain the overall effect of $\Sm$ to make the model
more fair. As an alternative, \citet{zafar} constrain the effects of the
individual sensitive attributes $S_i$: in a linear model (ZLM)
\begin{equation}
  \min_{\bbeta} \E\left[(\y - \Xm\bbeta)^2\right]
  \st |\COV(\Xm\bbeta, S_i)| < c, c \in \mathbb{R}^+
\label{eq:zlm}
\end{equation}
and in a logistic regression model (ZLRM)
\begin{equation}
  \max_{\bbeta} D(\bbeta)
  \st |\COV(\Xm\bbeta, S_i)| < c, c \in \mathbb{R}^+.
\label{eq:zlrm}
\end{equation}
Clearly, bounding the overall effect of $\Sm$ bounds the effects of the
individual $S_i$, but the former does not force the same bound on all $S_i$; and
$c$ is unbound which makes it more difficult to choose. Furthermore,
\citet{zafar} uses $\Xm$ instead $\wU$ as explanatory variables which leads to a
catastrophic loss of predictive accuracy for models that are constrained to be
almost perfectly fair ($c \to 0$) as demonstrated in \citet{stco21}. More in
general, this happens in any model that takes this approach because the
non-discriminating information in $\Xm$ is removed by the constraint together
with the discriminating information since the two are not separated and are
instead linked to the same regression coefficients. However, both ZLM and ZLRM
have the advantage that their model equations do not involve $\Sm$, which means
that we do not need to collect the sensitive attributes for the new observations
we want to predict. Hence they are a valid alternative to FRRM and FGRRM when
this is an issue.

\section{Features and Software Architecture}
\label{sec:features}

\begin{table}[t]
\centering
\begin{tabular}{lll}
  \hline
  \textbf{Function}                  & \textbf{Model} & \textbf{Reference} \\
  \hline
  \code{frrm()}, \code{fgrrm()}      & F(G)RRM  & \citet{stco21}    \\
  \code{nclm()}                      & NCLM     & \citet{komiyama}  \\
  \code{zlm()}, \code{zlm.orig()}    & ZLM      & \citet{zafar}     \\
  \code{zlrm()}, \code{zlrm.orig()}  & ZLRM     & \citet{zafar}     \\
  \hline
\end{tabular}
\caption{Fair models implemented in \pkg{fairml} and their canonical
  references.}
\label{tab:models}
\end{table}

The \pkg{fairml} package is centred around the functions implementing the models
discussed in the previous section: FRRM is implemented in \code{frrm()}, FGRRM
in \code{fgrrm()}, NCLM in \code{nclm()}, ZLM in \code{zlm()} and ZLRM in
\code{zlrm()}. All these functions are summarised in Table~\ref{tab:models} and
have similar signatures:
\begin{Schunk}
\begin{Sinput}
frrm(response, predictors, sensitive, unfairness, definition = "sp-komiyama",
  lambda = 0, save.auxiliary = FALSE)
fgrrm(response, predictors, sensitive, unfairness, definition = "sp-komiyama",
  family = "binomial", lambda = 0, save.auxiliary = FALSE)
nclm(response, predictors, sensitive, unfairness, covfun, lambda = 0,
  save.auxiliary = FALSE)
zlm(response, predictors, sensitive, unfairness)
zlrm(response, predictors, sensitive, unfairness)
\end{Sinput}
\end{Schunk}
In the above, \code{response} is the response variable (denoted $\y$ in Section
\ref{sec:fgrrm}), \code{predictors} are the non-sensitive explanatory variables
($\Xm$), \code{sensitive} are the sensitive attributes ($\Sm$) and
\code{unfairness} is the amount of unfairness allowed in the model ($r \in [0,
1]$). Both \code{predictors} and \code{sensitive} are internally transformed
into the respective design matrices. In the case of \code{zlm()} and
\code{zlrm()}, the covariance constraints in \mref{eq:zlm} and \mref{eq:zlrm}
are replaced with the corresponding correlation constraints to rescale them to
$[0, 1]$. As for the other arguments:
\begin{itemize}
  \item \code{save.auxiliary} controls whether the model $\wU = \Xm - \B^\T \Sm$
    that creates the decorrelated predictors in \mref{eq:frrm} (and implicitly
    in \mref{eq:nclm} and in \mref{eq:fgrrm}) is returned along with the main
    model.
  \item \code{definition} is a label that specifies which definition
    of fairness is used (Table~\ref{tab:fairness}):
    \begin{itemize}
      \item \code{"sp-komiyama"} is the fairness constraint in
        \mref{eq:nclm-bound}, which is also used in NCLM;
      \item \code{"eo-komiyama"} is the proportion of the variance or deviance
        of the fitted values explained by the sensitive attributes that is not
        explained by the original response;
      \item \code{"if-berk"} is the constraint in \mref{eq:individual},
        implemented and normalised as
          \begin{equation*}
            f(\balpha, \y, \Sm) = \frac{
              \sum\nolimits_{i, j} |y_i - y_j| (\si\balpha - \sj\balpha)^2
            }{
              \sum\nolimits_{i, j} |y_i - y_j| (\si\OLS{\balpha} - \sj\OLS{\balpha})^2
            } \in [0, 1],
          \end{equation*}
        where $\OLS{\balpha}$ are the coefficients associated with $\Sm$ in the
        model estimated without any fairness constraint (that is, $r = 1$).
    \end{itemize}
    As an alternative, users can provide a function implementing a custom
    fairness definition as illustrated in Section~\ref{sec:constraints}.
  \item \code{lambda} is an optional ridge penalty that is applied to $\bbeta$
    (in FRRM and FGRRM) or to $(\balpha, \bbeta)$ (in NCLM) to regularise the
    resulting models.
  \item \code{covfun} is the covariance function used internally by NCLM. It
    defaults to \code{cov()}; other options include the James-Stein estimator in
    \code{cov.shrink()} from \pkg{corpcor} \citep{corpcor} or a kernel estimator
    like that used in \citet{komiyama}.
\end{itemize}

\begin{table}[t]
\centering
\begin{tabular}{lll}
  \hline
  \textbf{Fairness}    & \textbf{Type}           & \textbf{Reference} \\
  \hline
  \code{"sp-komiyama"} & Statistical Parity      & \citet{komiyama}   \\
  \code{"eo-komiyama"} & Equality of Opportunity & \citet{stco21}     \\
  \code{"if-berk"}     & Individual Fairness     & \citet{berk}     \\
  \hline
\end{tabular}
\caption{Definitions of fairness implemented as constraints in \pkg{fairml}.}
\label{tab:fairness}
\end{table}

In addition, \pkg{fairml} provides two functions, \code{zlm.orig()} and
\code{zlrm.orig()}, that implement ZLM and ZLRM with the original constraint on
the covariances in \mref{eq:zlm} and \mref{eq:zlrm}. \code{frrm()} and
\code{fgrrm()} are built on the \pkg{glmnet} package
\citep{glmenet,coxenet,tayenet}; \code{zlm()} \code{zlrm()}, \code{zlm.orig()}
and \code{zlrm.orig()} are built on the \pkg{CVXR} package \citep{cvxr};
\code{nclm()} is built on the solver provided by the \pkg{cccp} package
\citep{cccp}.

All these functions return an object of class \code{"fair.model"} with an
additional class identifying the model estimator (\code{"frrm"}, \code{"fgrrm"},
\code{"nclm"}, \code{"zlm"}, or \code{"zlrm"}). These classes are used to
dispatch the models to the methods provided by the \pkg{stats} package for
built-in models:
\begin{itemize}
  \item \code{print()} and \code{summary()}, to print key facts about the model;
  \item \code{coef()}, \code{fitted()}, \code{residuals()} and \code{sigma()} to
    extract relevant parameters from the model;
  \item \code{deviance()}, \code{logLik()} and \code{nobs()} to assess the
    model's goodness of fit and to make \code{AIC()} and \code{BIC()} work;
  \item \code{plot()} to show diagnostic plots for model validation;
  \item \code{all.equal()} to compare two models;
  \item \code{predict()} to predict the values of the response for new
    observations.
\end{itemize}

In addition, \pkg{fairml} provides two tools for model selection. The first is
a function implementing cross-validation to evaluate a model's predictive
accuracy,
\begin{Schunk}
\begin{Sinput}
fairml.cv(response, predictors, sensitive, method = "k-fold", ..., unfairness,
  model, model.args = list(), cluster)
\end{Sinput}
\end{Schunk}
where:
\begin{itemize}
  \item \code{response}, \code{predictors}, \code{sensitive} and
    \code{unfairness} are the same as in the model estimation functions;
  \item \code{method} is the cross-validation scheme (\code{"k-fold"},
    \code{"hold-out"} or \code{"custom-folds"}) that will be used for resampling
    and \code{...} captures its optional arguments;
  \item \code{model} and \code{model.args} take the name of the model (as a
    string) and a list of optional arguments that will passed to it;
  \item \code{cluster} is an optional object created with the
    \code{makeCluster()} function from package \pkg{parallel} to enable parallel
    processing.
\end{itemize}

The second is a function that produces profile plots to evaluate how estimated
models change as a function of the fairness constraint,
\begin{Schunk}
\begin{Sinput}
fairness.profile.plot(response, predictors, sensitive, unfairness,
  legend = FALSE, type = "coefficients", model, model.args = list(), cluster)
\end{Sinput}
\end{Schunk}
where \code{type} determines what is plotted (\code{"coefficients"},
\code{"constraints"}, \code{"precision-recall"} or \code{"rmse"}) and
\code{legend} controls whether a legend with the variable names is displayed.
The remaining arguments have the same meaning as in \code{fairml.cv()}.

\begin{table}[t]
\centering \small
\begin{tabular}{p{0.35\linewidth}llp{0.20\linewidth}}
  \hline
  \textbf{Name}  & \textbf{Family} & \textbf{Response} & \textbf{Sensitive \newline attributes} \\
  \hline
  Adult \newline (\code{adult})
                 & Binomial        & Income            & Sex, race, age \\
  Bank \newline (\code{bank})
                 & Binomial        & Subscribed        & Age, marital status \\
  Communities \& crime \newline (\code{communities.and.crime})
                 & Gaussian        & Crime rate        & \% Black, \newline \% Foreign-born \\
  COMPAS \newline (\code{compas})
                 & Binomial        & Recidivism        & Sex, race, age \\
  Drug consumption \newline (\code{drug.consumption})
                 & Multinomial     & Consumption       & Age, gender, race \\
  Free light chain \newline (\code{flchain}, in \pkg{survival})
                 & Cox prop. haz.  & Days until death  & Age, sex \\
  Health and retirement \newline (\code{health.and.retirement})
                 & Poisson         & Score             & Marriage, gender, race, age \\
  Law school admissions \newline (\code{law.school.admissions})
                 & Gaussian        & GPA               & Age, race \\
  National longitudinal survey \newline (\code{national.longitudinal.survey})
                 & Gaussian        & Income            & Gender, age \\
  Obesity level \newline (\code{obesity.levels})
                 & Multinomial     & Obesity level     & Gender, age \\
  \hline
\end{tabular}
\caption{Real-world data sets used to benchmark fair machine learning models
  in the literature that are included \pkg{fairml}; \code{flchain} is actually
  provided by the \pkg{survey} package but its use in fair modelling is
  documented in \pkg{fairml}. For each data set, we report the name (including
  the name of the \proglang{R} object), the GLM family of the fair model,
  the response variable and the sensitive attributes.}
\label{tab:datasets}
\end{table}

Finally, \pkg{fairml} ships with a comprehensive collection of real-world data
sets used in the literature (Table~\ref{tab:datasets}): many of them come from
the UCI Machine Learning Repository \citep{uci}. This collection is provided to
make it easier to explore and benchmark fair machine learning models, including
those in \pkg{fairml}, as well as to serve as a basis for model comparisons in
future literature. All data sets have been minimally preprocessed and cleaned to
preserve their original features, as documented in the respective manual pages.

\section{Examples of Fair Modelling}
\label{sec:showcase}

In this last section we will use the \code{drug.consumption} and the
\code{national.longitudinal.survey} data sets from Table~\ref{tab:datasets} to
illustrate how to use \pkg{fairml} to select and estimate fair models.

\subsection{Drug Consumption}
\label{sec:drugs}

The drug consumption data set, originally studied in \citet{fehrman}, originates
from an online survey to evaluate an individual's risk of drug consumption with
respect to personality traits, impulsivity (\code{Impulsivity}), sensation
seeking (\code{SS}) and demographic information. Personality traits were
modelled using the ``five-factor model'' comprising scores for neuroticism
(\code{Nscore}), extroversion (\code{Escore}), openness to experience
(\code{Oscore}), agreeableness (\code{Ascore}) and conscientiousness
(\code{Cscore}). As for the demographic information, the survey recorded the age
(\code{Age}), gender (\code{Gender}), race (\code{Race}) and education level
(\code{Education}) of each respondent. The data set contains information on the
consumption of 18 psychoactive drugs including amphetamines, cocaine, crack,
ecstasy, heroin, ketamine, and others. In this example we will concentrate on
LSD use.

The respondents are a self-selected sample: they self-enrol during the
recruitment period. Therefore, it is natural to ask the question: \emph{is the
sample biased because individuals fail to enrol, or enrol and then refuse to
answer certain questions, because of their age, gender or race?} This form of
sampling bias has also been studied in the field of survey sampling, and how to
use sampling weights to remove it has been thoroughly studied in the literature
\citep[see, for instance,][]{lohr}.

After loading the data,
\begin{Schunk}
\begin{Sinput}
R> data(drug.consumption)
\end{Sinput}
\end{Schunk}
and extracting the response variable (\code{r}), the sensitive attributes
(\code{s}) and the predictors (\code{p}),
\begin{Schunk}
\begin{Sinput}
R> r = drug.consumption[, "LSD"]
R> s = drug.consumption[, c("Age", "Gender", "Race")]
R> p = drug.consumption[, c("Education", "Nscore", "Escore", "Oscore",
+                           "Ascore", "Cscore", "Impulsive", "SS")]
\end{Sinput}
\end{Schunk}
we merge levels with a low number of samples both in \code{Education} and in the
response.
\begin{Schunk}
\begin{Sinput}
R> levels(p$Education) =
+    c("at.most.18y", "at.most.18y", "at.most.18y", "at.most.18y",
+      "university", "diploma", "bachelor", "master", "phd")
R> levels(r) = c("never", ">=1y", ">=1y", "<1y", "<1m", "<1m", "<1m")
\end{Sinput}
\end{Schunk}
As a result, all the levels of these two variables are observed in at least
89 samples out of 1885
and most are observed in more than 200 samples.

We can then fit a multinomial FGRRM model from \code{r}, \code{s} and \code{p}
with a minimal amount of unfairness ($r = 0.05$) and a small additional ridge
penalty for smoothing.

\begin{Schunk}
\begin{Sinput}
R> m = fgrrm(response = r, sensitive = s, predictors = p,
+        family = "multinomial", unfairness = 0.05, lambda = 0.1)
R> summary(m)
\end{Sinput}
\begin{Soutput}

Fair Linear Regression Model

Method: Fair Generalized Ridge Regression

Call:
fgrrm(response = r, predictors = p, sensitive = s, unfairness = 0.05,
    family = "multinomial", lambda = 0.1)

Coefficients:
                       never       >=1y        <1y         <1m
(Intercept)             1.1079621   0.1549633  -0.4910178  -0.7719076
Age25.34                0.0044356   0.0002583  -0.0019688  -0.0027251
Age35.44                0.0031367   0.0170390  -0.0104855  -0.0096902
Age45.54                0.0052437   0.0149946  -0.0117205  -0.0085178
Age55.64                0.0026464   0.0192400  -0.0123804  -0.0095060
Age65.                  0.0326250  -0.0116477  -0.0116190  -0.0093583
 [ reached getOption("max.print") -- omitted 19 rows ]

Ridge penalty (sensitive attributes): 9.601 (predictors): 0.1
Log-likelihood: -1944
Komiyama's R^2 (statistical parity): 0.05 with bound: 0.05
\end{Soutput}
\end{Schunk}
The model summary contains the sets of coefficients that the multinomial
logistic regression fits for each level of the response;\footnote{The
\code{multinom()} function in \pkg{nnet} returns the coefficients for all but
the first level of the response, after subtracting the corresponding
coefficients for the first level. \pkg{glmnet} returns the raw coefficients
for all levels and \pkg{fairml} follows the same convention.} the overall
ridge penalties applied to the sensitive attributes and to the predictors; and
confirms that the desired level of (un)fairness has been achieved.

Is this model a good fit for the data? The diagnostic plots generated by
\begin{Schunk}
\begin{Sinput}
R> plot(m, support = TRUE)
\end{Sinput}
\begin{figure}[t!]

{\centering \includegraphics[width=0.8\linewidth,height=0.8\linewidth]{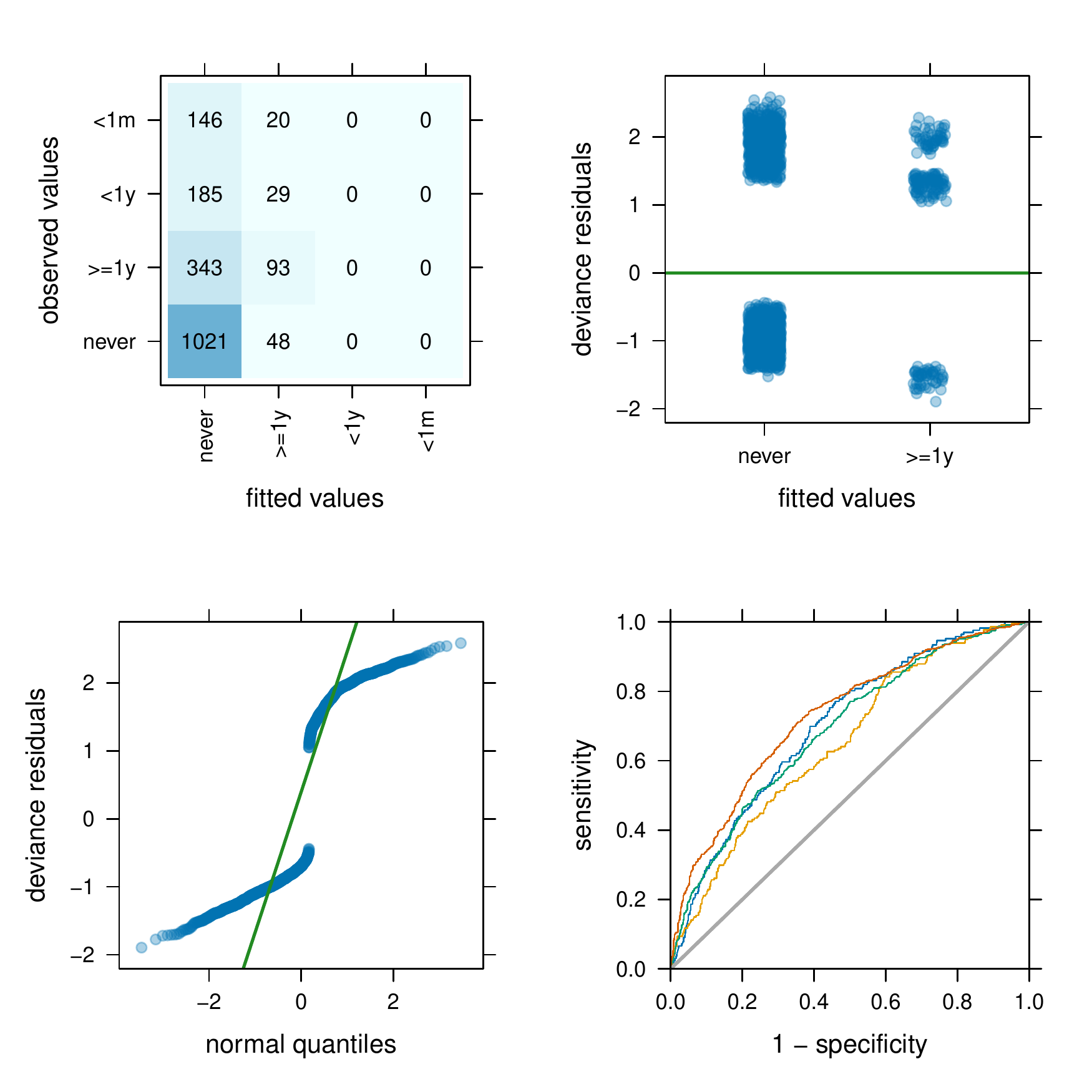}

}

\caption[Diagnostic plots produced by \code{plot(m)}]{Diagnostic plots produced by \code{plot(m)}: a confusion matrix (top left), deviance residuals against fitted.values (top right), qq-plot of the deviance residuals (bottom left), ROC curves for the four levels of the response variable.}\label{fig:diagnostic-plots}
\end{figure}
\end{Schunk}
and shown in Figure~\ref{fig:diagnostic-plots} suggest that the model is
impacted by the imbalance between the number of people who have never used LSD
and those who have (top-left panel). The ROC curves , computed one-versus-rest
for each level of the response, are all similar and suggest that the model is an
acceptable fit for the data (bottom-right panel). Running the standard 10 runs
of 10-fold cross-validation suggested by \citet{elemstatlearn}, we confirm that
neither precision nor recall are particularly high.

\begin{Schunk}
\begin{Sinput}
R> fairml.cv(response = r, sensitive = s, predictors = p, model = "fgrrm",
+    unfairness = 0.05, method = "k-fold", k = 10, runs = 10,
+    model.args = list(family = "multinomial", lambda = 0.1))
\end{Sinput}
\begin{Soutput}

  k-fold cross-validation for fair models

  model:
                                      Fair Generalized Ridge Regression
  number of folds:                       10
  number of runs:                        10
  average loss over the runs:
    precision:                           0.6443056
    recall:                              0.2144495
  standard deviation of the loss:
    precision:                           0.009535637
    recall:                              0.006509704
\end{Soutput}
\end{Schunk}

\pagebreak

This leaves us with two questions:
\begin{enumerate}
  \item Does the fairness constraint have a strong impact on the goodness of fit
    and on the predictive performance of the model?
  \item Would merging the levels of the response variable into a binary
    ``used''/``never used'' variable improve the model by making the data
    balanced?
\end{enumerate}
We explore the answer to the first questions by plotting the profile of the
estimated regression coefficients against the fairness constraint in
Figure~\ref{fig:profile-plot}.
\begin{Schunk}
\begin{Sinput}
R> fairness.profile.plot(response = r, sensitive = s, predictors = p,
+      model = "fgrrm", type = "coefficients",
+      model.args = list(family = "multinomial", lambda = 0.1))
\end{Sinput}
\begin{figure}[t!]

{\centering \includegraphics[width=0.8\linewidth,height=0.6\linewidth]{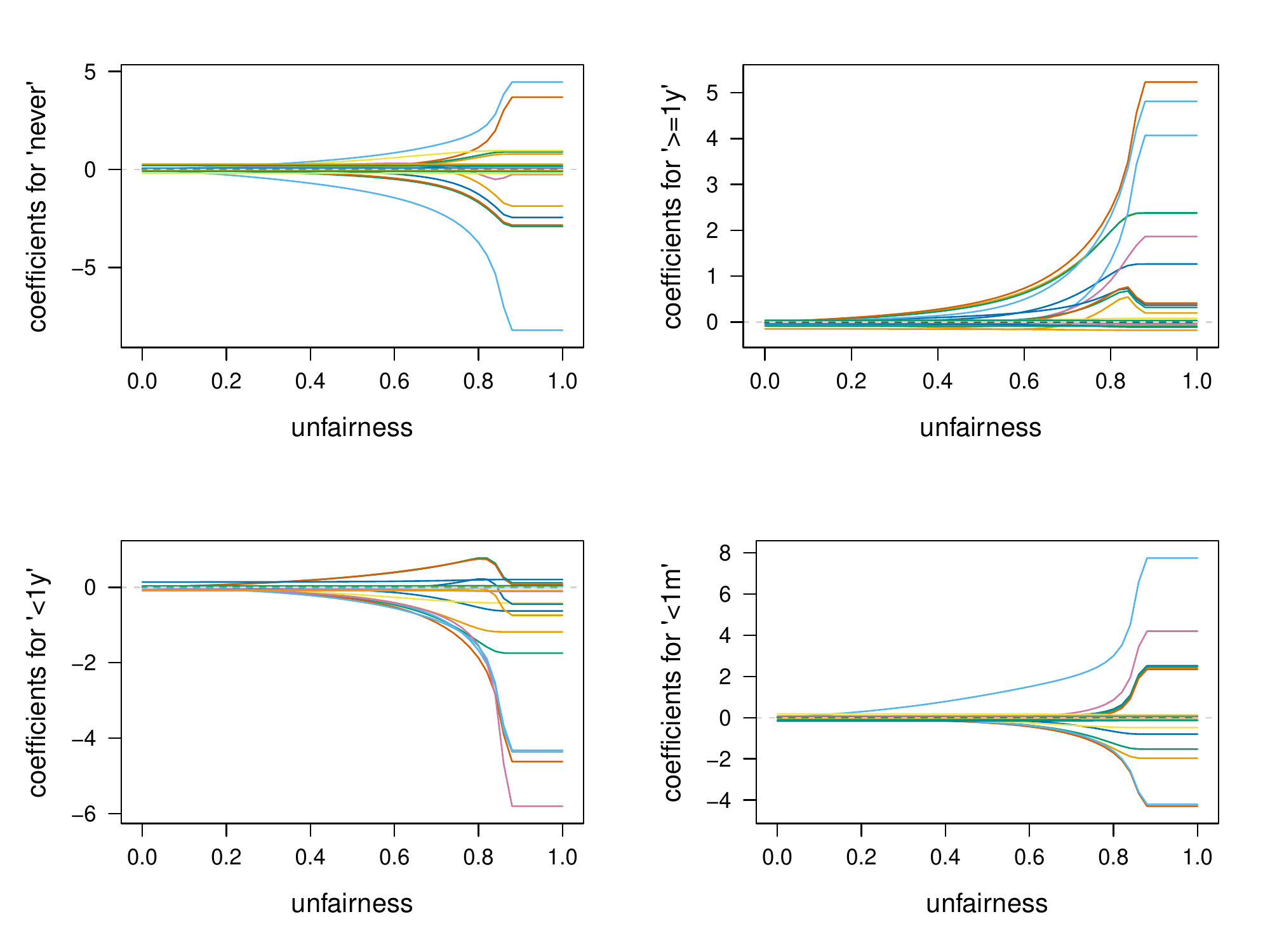}

}

\caption[Profile plots for the regression coefficients as a function of the fairness constraint for all the levels of the response variable]{Profile plots for the regression coefficients as a function of the fairness constraint for all the levels of the response variable.}\label{fig:profile-plot}
\end{figure}
\end{Schunk}
All regression coefficients are markedly shrunk towards zero for small values of
\code{unfairness}: they take much larger values (in absolute value) when the
model is unrestricted at \code{unfairness} equal to \code{1}. This suggests that
\code{Age}, \code{Gender} and \code{Race} are strongly associated with the
predictors, in addition to their direct effect on the response. Their profiles
flatten around \code{unfairness} equal to \code{0.85} which means that
approximately 85\% of the deviance of the model would be explained by the
sensitive attributes when the fairness constraint becomes inactive.

As for re-balancing the response by transforming it into a binary variable,
\begin{Schunk}
\begin{Sinput}
R> levels(r) = c("never", "used", "used", "used")
R> table(r)
\end{Sinput}
\begin{Soutput}
r
never  used
 1069   816
\end{Soutput}
\end{Schunk}
testing the predictive performance of the (no longer multinomial) FGRRM logistic
regression model with cross-validation shows a sharp increase in recall and
similar precision.
\begin{Schunk}
\begin{Sinput}
R> fairml.cv(response = r, sensitive = s, predictors = p, model = "fgrrm",
+    unfairness = 0.05, method = "k-fold", k = 10, runs = 10,
+    model.args = list(family = "binomial", lambda = 0.1))
\end{Sinput}
\begin{Soutput}

  k-fold cross-validation for fair models

  model:
                                      Fair Generalized Ridge Regression
  number of folds:                       10
  number of runs:                        10
  average loss over the runs:
    precision:                           0.6866465
    recall:                              0.5515931
  standard deviation of the loss:
    precision:                           0.002800798
    recall:                              0.007156275
\end{Soutput}
\end{Schunk}
Therefore, we can conclude that the imbalance in the original response had a
definite impact on the predictive accuracy of the multinomial FGRRM model.

\subsection{Income and Labour Market}
\label{sec:nlsy}

The National Longitudinal Survey data set contains the results of a survey from
the U.S. Bureau of Labor Statistics to gather information on labour market
activities \citep{nls}. Along with the incomes in 1990, 1996 and 2006
(\code{income90}, \code{income96} and \code{income06}), the survey records the
gender (\code{gender}), age (\code{age}) and race (\code{race}) of the
respondents (our sensitive attributes), their physical characteristics (height,
weight, general health), their criminal records (number of illegal acts and
charges) and their level of education (\code{grade90}). As was the case in
Section~\ref{sec:drugs}, we want to remove the bias introduced by the sensitive
attributes through sampling bias and other mechanisms.

After loading the data, merging the levels of \code{race} and \code{grade90}
with low numbers of observations, and separating the response variable
(\code{income90} in this example), the predictors and the sensitive attributes,
\begin{Schunk}
\begin{Sinput}
R> data(national.longitudinal.survey)
R> nlsy = national.longitudinal.survey
R> levels(nlsy$grade90)[1:5] = "7TH GRADE"
R> levels(nlsy$race)[c(3, 14, 15)] = "CHINESE"
R> r = nlsy[, "income90"]
R> s = nlsy[, c("gender", "age", "race")]
R> p = nlsy[, setdiff(names(nlsy),
+               c("income90", "income96", "income06", "gender", "race", "age"))]
\end{Sinput}
\end{Schunk}
we can fit an FRRM model (or equivalently a Gaussian FGRRM) as is commonly done
in the literature \citep[see, for instance,][]{komiyama}.

\begin{Schunk}
\begin{Sinput}
R> m = frrm(response = r, sensitive = s, predictors = p, unfairness = 0.05)
R> summary(m)
\end{Sinput}
\begin{Soutput}

Fair Linear Regression Model

Method: Fair Ridge Regression

Call:
frrm(response = r, predictors = p, sensitive = s, unfairness = 0.05)

Coefficients:
 (Intercept)  genderFemale           age     raceBLACK   raceCHINESE
   1.6102737    -0.1873255     0.0186816    -0.0980725     0.0765317
 [ reached getOption("max.print") -- omitted 62 entries ]

Ridge penalty (sensitive attributes): 4.081 (predictors): 0
Log-likelihood: -7835
Residual standard error: 1.202
Multiple R^2: 0.2337
Komiyama's R^2 (statistical parity): 0.05 with bound: 0.05
\end{Soutput}
\end{Schunk}
The $R^2$ coefficient reported by \code{summary()} suggests that the model is
not a good fit for the data. Is this caused by the fairness constraint?
Comparing the proportions of variance explained by the sensitive attributes and
by the other predictors with \code{fairness.profile.plot()}, we can see in
Figure~\ref{fig:fit-profile-plot} that without the fairness constraint (which
becomes inactive at $r \approx 0.45$) the sensitive attributes explain nearly as
much variability as the predictors.

\begin{Schunk}
\begin{Sinput}
R> fairness.profile.plot(response = r, sensitive = s, predictors = p,
+      model = "frrm", type = "constraints", legend = TRUE)
\end{Sinput}
\begin{figure}[t!]

{\centering \includegraphics[width=0.8\linewidth,height=0.4\linewidth]{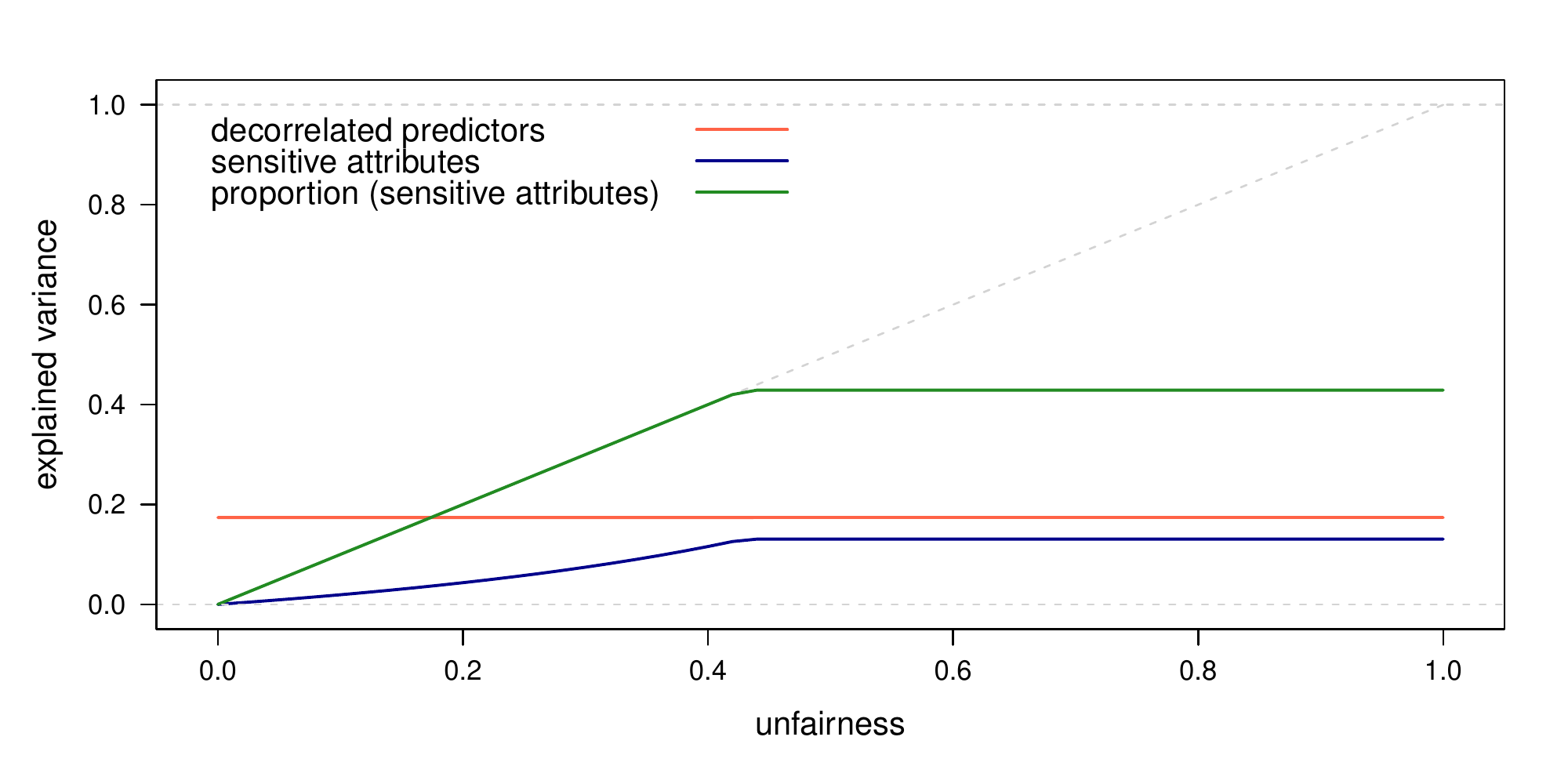}

}

\caption[Profile plot for the variance proportions explained by the decorrelated predictors and the sensitive attributes]{Profile plot for the variance proportions explained by the decorrelated predictors and the sensitive attributes.}\label{fig:fit-profile-plot}
\end{figure}
\end{Schunk}

While removing or relaxing the fairness constraint would nearly double the $R^2$
of the model, the diagnostic plots produced by \code{plot()}
(Figure~\ref{fig:frrm-diagnostic-plots}) highlight two underlying issues with
the data: higher incomes are truncated to $7.428$, which impacts the residuals
in the right tail of the qq-plot and creates a pattern of points in the first
two plots; and the response is bound below by zero.

\begin{Schunk}
\begin{figure}[t!]

{\centering \includegraphics[width=\linewidth,height=0.4\linewidth]{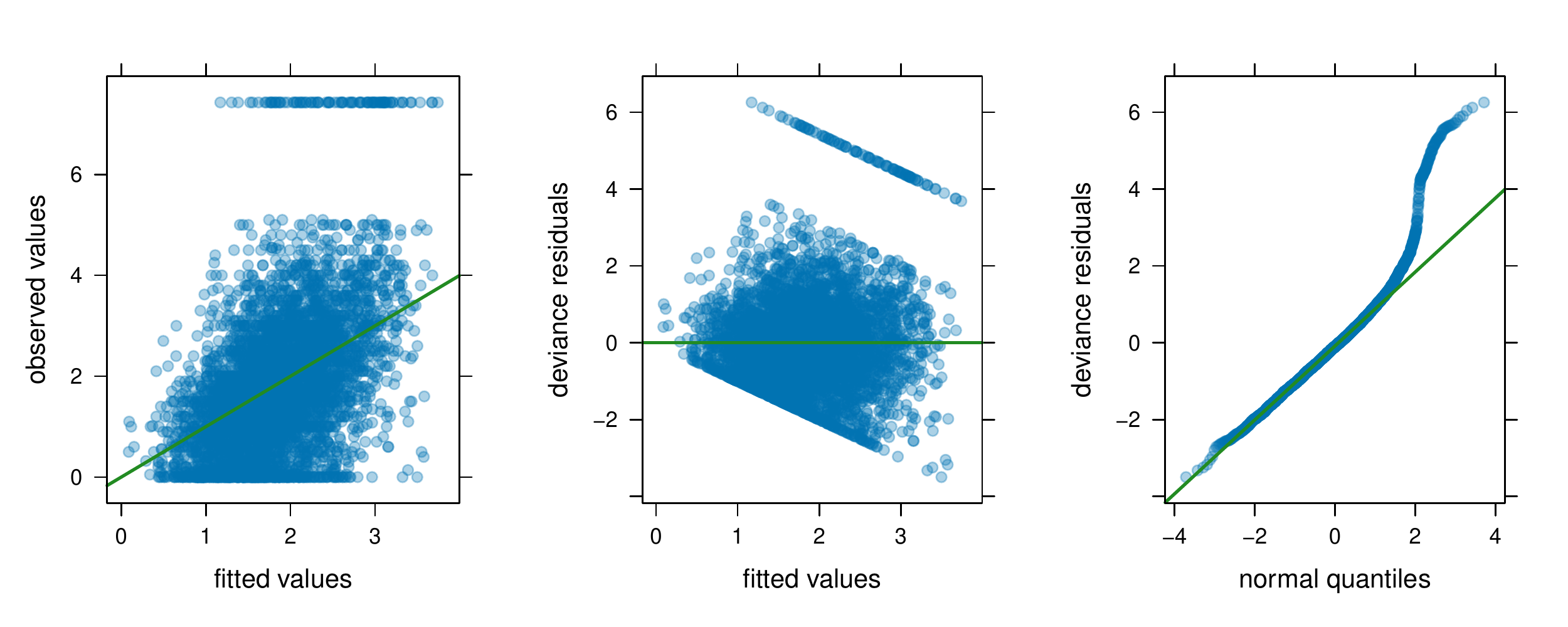}

}

\caption[Diagnostic plots produced by \code{plot()} for an FGRRM model with a continuous response]{Diagnostic plots produced by \code{plot()} for an FGRRM model with a continuous response: observed versus fitted values (left), deviance residuals versus fitted values (centre) and a qq-plot of the deviance residuals (right).}\label{fig:frrm-diagnostic-plots}
\end{figure}
\end{Schunk}

To address them, we may consider removing the truncated observations and fitting
a Poisson FGRRM model on the data, which is a natural choice for a non-negative
response.
\begin{Schunk}
\begin{Sinput}
R> large.income = (r == max(r))
R> table(large.income)
\end{Sinput}
\begin{Soutput}
large.income
FALSE  TRUE
 4816    92
\end{Soutput}
\begin{Sinput}
R> m = fgrrm(response = r[!large.income], sensitive = s[!large.income, ],
+        predictors = p[!large.income, ], family = "poisson",
+        unfairness = 0.05)
\end{Sinput}
\end{Schunk}

However, comparing the two models using cross-validation on the same data (and
using the same folds) reveals that the original FRRM model appears to have better
predictive accuracy for the chosen level of fairness than the Poisson FGRRM.
\begin{Schunk}
\begin{Sinput}
R> xval.lm =
+    fairml.cv(response = r[!large.income], sensitive = s[!large.income, ],
+      predictors = p[!large.income, ], model = "frrm", unfairness = 0.05,
+      method = "k-fold", k = 10, runs = 10)
R> xval.glm =
+    fairml.cv(response = r[!large.income], sensitive = s[!large.income, ],
+      predictors = p[!large.income, ], model = "fgrrm", unfairness = 0.05,
+      model.args = list(family = "poisson"), method = "custom-folds",
+      folds = cv.folds(xval.lm))
R> summary(cv.loss(xval.lm))
\end{Sinput}
\begin{Soutput}
   Min. 1st Qu.  Median    Mean 3rd Qu.    Max.
 0.9908  0.9929  0.9937  0.9938  0.9950  0.9967
\end{Soutput}
\begin{Sinput}
R> summary(cv.loss(xval.glm))
\end{Sinput}
\begin{Soutput}
   Min. 1st Qu.  Median    Mean 3rd Qu.    Max.
  1.128   1.130   1.131   1.131   1.133   1.134
\end{Soutput}
\end{Schunk}

The other two models implemented in \pkg{fairml}, \code{zlm()} from
\citet{zafar} and \code{nclm()} from \citet{komiyama}, also appear to fit the
data less well than FRRM.

\begin{Schunk}
\begin{Sinput}
R> m2 = nclm(response = r[!large.income], sensitive = s[!large.income, ],
+         predictors = p[!large.income, ], unfairness = 0.05)
\end{Sinput}
\begin{Soutput}
Loading required namespace: cccp
\end{Soutput}
\begin{Sinput}
R> m3 = zlm(response = r[!large.income], sensitive = s[!large.income, ],
+         predictors = p[!large.income, ], unfairness = 0.05)
\end{Sinput}
\begin{Soutput}
Loading required namespace: CVXR
\end{Soutput}
\begin{Sinput}
R> summary(m2)
\end{Sinput}
\begin{Soutput}

Fair Linear Regression Model

Method: Komiyama et al. (2018)

Call:
nclm(response = r[!large.income], predictors = p[!large.income,
    ], sensitive = s[!large.income, ], unfairness = 0.05)

Coefficients:
 (Intercept)  genderFemale           age     raceBLACK   raceCHINESE
   1.4635456    -0.1974729     0.0162676     0.0076029     0.2002531
 [ reached getOption("max.print") -- omitted 62 entries ]

Ridge penalty: 0
Custom covariance matrix: FALSE
Log-likelihood: -6757
Residual standard error: 0.9912
Multiple R^2: 0.248
Komiyama's R^2 (statistical parity): 0.04984 with bound: 0.05
\end{Soutput}
\begin{Sinput}
R> summary(m3)
\end{Sinput}
\begin{Soutput}

Fair Linear Regression Model

Method: Zafar's Linear Regression

Call:
zlm(response = r[!large.income], predictors = p[!large.income,
    ], sensitive = s[!large.income, ], unfairness = 0.05)

Coefficients:
(Intercept)    grade90.L    grade90.Q    grade90.C    grade90.4
  1.4581758    1.3300355    0.0028437   -0.2713558    0.0125724
 [ reached getOption("max.print") -- omitted 34 entries ]

Log-likelihood: -6940
Residual standard error: 1.027
Multiple R^2: 0.1887
Marginal correlation (disparate impact):
genderFemale          age    raceBLACK  raceCHINESE  raceENGLISH
    0.032564     0.007244     0.036622     0.013714     0.041502
 [ reached getOption("max.print") -- omitted 23 entries ]
with bound: 0.05
\end{Soutput}
\end{Schunk}

Therefore, we can conclude that FRRM fits the data best at
\code{unfairness = 0.05} among the models we have examined.

\subsection{Using Different Fairness Constraints}
\label{sec:constraints}

The last feature of \pkg{fairml} we will showcase is the ability to plug
custom mathematical characterisations of fairness into \code{frrm()} and
\code{fgrrm()}. Built-in fairness definitions are identified by the labels
listed in Table~\ref{tab:fairness}. Custom definitions can be provided as
functions with signature \code{function(model, y, S, U, family)} where
\code{model} is the model whose fairness we are evaluating, \code{family} is the
GLM family the model belongs to, and \code{y, S, U} are $\y, \Sm, \U$ from
\mref{eq:frrm2}. With the help of \code{str()} and a dummy function, we can see
that the \code{model} object contains several key quantities describing the
model we are evaluating.

\begin{Schunk}
\begin{Sinput}
R> dummy.fairness =
+    function(model, y, S, U, family) { str(model); return(c(value = 0)) }
R>
R> m = fgrrm(response = r, sensitive = s, predictors = p, unfairness = 0.05,
+        definition = dummy.fairness, family = "gaussian")
\end{Sinput}
\begin{Soutput}
List of 5
 $ coefficients: Named num [1:67] 0.6526 -0.7423 0.0754 -0.0635 0.4791 ...
  ..- attr(*, "names")= chr [1:67] "(Intercept)" "genderFemale" "age" ...
 $ deviance    : num 6350
 $ loglik      :Class 'logLik' : -7596 (df=68)
 $ fitted      : num [1:4908] 1.86 2.92 1.63 2.32 3.5 ...
 $ residuals   : num [1:4908] 0.1419 -0.6214 0.0683 0.6846 0.5017 ...
\end{Soutput}
\end{Schunk}

The only requirement for this function is that it should return an array with an
element named \code{"value"} that takes a value between 0 (perfect fairness) and
1 (no constraint). Built-in functions return additional elements that are then
used in \linebreak \code{fairness.profile.plot()}, but any element other than
that named \code{"value"} will be ignored for custom fairness definitions. So,
for instance, we can define fairness as a weighted mean of statistical parity
and individual fairness like \citet{berk} did by reusing the internal functions
that \pkg{fairml} uses to estimate them.
\begin{Schunk}
\begin{Sinput}
R> combined.fairness = function(model, y, S, U, family) {
+
+    value =
+      0.4 * fairml:::fgrrm.sp.komiyama(model, y, S, U, family)[["value"]] +
+      0.6 * fairml:::fgrrm.if.berk(model, y, S, U, family)[["value"]]
+
+    return(c(value = value))
+
+  }#COMBINED.FAIRNESS
\end{Sinput}
\end{Schunk}

Or we can implement a completely different definition of fairness, such as a
bound on the correlation between fitted values and sensitive attributes in the
spirit of \citet{zafar}.
\begin{Schunk}
\begin{Sinput}
R> custom.fairness =
+    function(model, y, S, U, family)
+      return(c(value = max(abs(cor(model$fitted, S)))))
\end{Sinput}
\end{Schunk}

Or, assuming we only have a single binary sensitive attribute in the data, we
can bound the p-value of a Kolmogorov-Smirnov test between the distributions of
the fitted values for the two groups identified by the sensitive attribute.
\begin{Schunk}
\begin{Sinput}
R> custom.fairness = function(model, y, S, U, family) {
+
+    group1 = model$fitted[S == 0]
+    group2 = model$fitted[S == 1]
+    value = ks.test(group1, group2)$p.value
+
+    return(c(value = value))
+
+  }#CUSTOM.FAIRNESS
\end{Sinput}
\end{Schunk}

In essence, the information provided in the arguments to the custom function
combined with the functionality of other \proglang{R} packages allows for a
great deal of flexibility.

\pagebreak

\section{Summary and Discussion}
\label{sec:discussion}

Algorithmic fairness is a topic with important practical applications and a
quickly-evolving research field. The core of the \pkg{fairml} package is our
previous work in \citet{stco21}, which separates the estimation of a fair linear
model from the mathematical characterisation of fairness. On the one hand, this
modular design choice allows for a compact implementation that supports all GLM
families which makes \pkg{fairml} useful in a variety of application fields. On
the other hand, the ability to plug any definition of fairness in the model
without having to reimplement model estimation as well is a valuable asset in
research. Furthermore, \pkg{fairml} produces models whose statistical properties
and best practices are well known (as opposed to the black-box estimators based
on numerical optimisers that make up most of the literature) and provides the
standard tools to validate and evaluate them.


\begin{thebibliography}{53}
\newcommand{\enquote}[1]{``#1''}
\providecommand{\natexlab}[1]{#1}
\providecommand{\url}[1]{\texttt{#1}}
\providecommand{\urlprefix}{URL }
\expandafter\ifx\csname urlstyle\endcsname\relax
  \providecommand{\doi}[1]{doi:\discretionary{}{}{}#1}\else
  \providecommand{\doi}{doi:\discretionary{}{}{}\begingroup
  \urlstyle{rm}\Url}\fi
\providecommand{\eprint}[2][]{\url{#2}}

\bibitem[{Agarwal \emph{et~al.}(2018)Agarwal, Beygelzimer, Dudik
  \emph{et~al.}}]{reductions}
Agarwal A, Beygelzimer A, Dudik M, \emph{et~al.} (2018).
\newblock \enquote{{A Reductions Approach to Fair Classification}.}
\newblock \emph{Proceedings of Machine Learning Research}, \textbf{80}, 60--69.
\newblock {35th International Conference on Machine Learning (ICML)}.

\bibitem[{Agarwal \emph{et~al.}(2019)Agarwal, Dudik, and Wu}]{grouploss}
Agarwal A, Dudik M, Wu ZS (2019).
\newblock \enquote{{Fair Regression: Quantitative Definitions and
  Reduction-Based Algorithms}.}
\newblock \emph{Proceedings of Machine Learning Research}, \textbf{97},
  120--129.
\newblock {36th International Conference on Machine Learning}.

\bibitem[{Angwin \emph{et~al.}(2016)Angwin, Larson, Mattu, and
  Kirchner}]{compas}
Angwin J, Larson J, Mattu S, Kirchner L (2016).
\newblock \emph{{Machine Bias: There's Software Used Across the Country to
  Predict Future Criminals. And It's Biased Against Blacks}}.
\newblock ProPublica.
\newblock
  \urlprefix\url{https://www.propublica.org/article/machine-bias-riskassessments-in-criminal-sentencing}.

\bibitem[{{Barredo Arrieta} \emph{et~al.}(2020){Barredo Arrieta},
  {D{\'i}az-Rodr{\'i}guez}, {Del Ser} \emph{et~al.}}]{xai}
{Barredo Arrieta} A, {D{\'i}az-Rodr{\'i}guez} N, {Del Ser} J, \emph{et~al.}
  (2020).
\newblock \enquote{{Explainable Artificial Intelligence (XAI): Concepts,
  Taxonomies, Opportunities and Challenges Toward Responsible AI}.}
\newblock \emph{Information Fusion}, \textbf{58}, 82--115.

\bibitem[{Bellamy \emph{et~al.}(2019)Bellamy, Dey, Hind \emph{et~al.}}]{ai360}
Bellamy RKE, Dey K, Hind M, \emph{et~al.} (2019).
\newblock \enquote{{AI Fairness 360: An Extensible Toolkit for Detecting and
  Mitigating Algorithmic Bias}.}
\newblock \emph{IBM Journal of Research and Development}, \textbf{63}(4/5),
  1--15.
\newblock \doi{10.1147/JRD.2019.2942287}.

\bibitem[{Berk \emph{et~al.}(2017)Berk, Heidari, Jabbari \emph{et~al.}}]{berk}
Berk R, Heidari H, Jabbari S, \emph{et~al.} (2017).
\newblock \enquote{{A Convex Framework for Fair Regression}.}
\newblock In \emph{{Fairness, Accountability, and Transparency in Machine
  Learning (FATML)}}.

\bibitem[{Berk \emph{et~al.}(2021)Berk, Heidari, Jabbari
  \emph{et~al.}}]{heidari}
Berk R, Heidari H, Jabbari S, \emph{et~al.} (2021).
\newblock \enquote{{Fairness in Criminal Justice Risk Assessments: The State of
  the Art}.}
\newblock \emph{Sociological Methods \& Research}, \textbf{50}(1), 3--44.
\newblock \doi{10.1177/0049124118782533}.

\bibitem[{Calders \emph{et~al.}(2013)Calders, Karim, Kamiran
  \emph{et~al.}}]{calders}
Calders T, Karim A, Kamiran F, \emph{et~al.} (2013).
\newblock \enquote{{Controlling Attribute Effect in Linear Regression}.}
\newblock In \emph{{Proceedings of the 13th IEEE International Conference on
  Data Mining}}, pp. 71--80.
\newblock \doi{10.1109/ICDM.2013.114}.

\bibitem[{Calmon \emph{et~al.}(2017)Calmon, Wei, Vinzamuri
  \emph{et~al.}}]{calmon}
Calmon F, Wei D, Vinzamuri B, \emph{et~al.} (2017).
\newblock \enquote{{Optimized Pre-Processing for Discrimination Prevention}.}
\newblock In \emph{Advances in Neural Information Processing Systems},
  volume~30, pp. 3992--4001.

\bibitem[{Cath \emph{et~al.}(2018)Cath, Wachter, Mittelstadt
  \emph{et~al.}}]{cath}
Cath C, Wachter S, Mittelstadt B, \emph{et~al.} (2018).
\newblock \enquote{{Artificial Intelligence and the `Good Society': the US, EU,
  and UK Approach}.}
\newblock \emph{Science and Engineering Ethics}, \textbf{24}(2), 505--528.
\newblock \doi{10.1007/s11948-017-9901-7}.

\bibitem[{Chora{\'s} \emph{et~al.}(2020)Chora{\'s}, Pawlicki, Puchalski, and
  Kozik}]{choras}
Chora{\'s} M, Pawlicki M, Puchalski D, Kozik R (2020).
\newblock \enquote{{Machine Learning---The Results Are Not the only Thing that
  Matters! What About Security, Explainability and Fairness?}}
\newblock In \emph{{Proceedings of the International Conference on
  Computational Science (ICCS)}}, pp. 615--628.
\newblock \doi{10.1007/978-3-030-50423-6_46}.

\bibitem[{Chzhen \emph{et~al.}(2020)Chzhen, Denis, Hebiri
  \emph{et~al.}}]{chzhen}
Chzhen E, Denis C, Hebiri M, \emph{et~al.} (2020).
\newblock \enquote{{Fair Regression via Plug-In Estimator and Recalibration
  with Statistical Guarantees}.}
\newblock In \emph{{Advances in Neural Information Processing Systems}},
  volume~33, pp. 19137--19148.

\bibitem[{Cotter \emph{et~al.}(2019)Cotter, Gupta, Jiang
  \emph{et~al.}}]{cotter}
Cotter A, Gupta M, Jiang H, \emph{et~al.} (2019).
\newblock \enquote{{Training Well-Generalizing Classifiers for Fairness Metrics
  and Other Data-Dependent Constraints}.}
\newblock \emph{Proceedings of Machine Learning Research}, \textbf{97},
  1397--1405.
\newblock 36th International Conference on Machine Learning.

\bibitem[{{D'Alessandro} \emph{et~al.}(2017){D'Alessandro}, {O'Neil}, and
  LaGatta}]{lagatta}
{D'Alessandro} B, {O'Neil} C, LaGatta T (2017).
\newblock \enquote{{Conscientious Classification: A Data Scientist's Guide to
  Discrimination-Aware Classification}.}
\newblock \emph{Big Data}, \textbf{5}(2), 120--134.
\newblock \doi{10.1089/big.2016.0048}.

\bibitem[{de~B.~G.~{de Oliveira} \emph{et~al.}(2021)de~B.~G.~{de Oliveira},
  {Vieira}, Silva \emph{et~al.}}]{predfairness}
de~B~G~{de Oliveira} T, {Vieira} LP, Silva GRL, \emph{et~al.} (2021).
\newblock \emph{{\pkg{predfairness:} Discrimination Mitigation for Machine
  Learning Models}}.
\newblock \urlprefix\url{https://cran.r-project.org/package=predfairness}.

\bibitem[{{Del Barrio} \emph{et~al.}(2020){Del Barrio}, Gordaliza, and
  Loubes}]{delbarrio}
{Del Barrio} E, Gordaliza P, Loubes JM (2020).
\newblock \emph{{Review of Mathematical Frameworks for Fairness in Machine
  Learning}}.
\newblock \urlprefix\url{https://arxiv.org/abs/2005.13755}.

\bibitem[{D{\'i}az \emph{et~al.}(2019)D{\'i}az, Johnson, Lazar
  \emph{et~al.}}]{diaz}
D{\'i}az M, Johnson I, Lazar A, \emph{et~al.} (2019).
\newblock \enquote{{Addressing Age-Related Bias in Sentiment Analysis}.}
\newblock In \emph{{Proceedings of the Twenty-Eighth International Joint
  Conference on Artificial Intelligence}}, pp. 6146--6150.
\newblock \doi{10.24963/ijcai.2019/852}.

\bibitem[{Do \emph{et~al.}(2022)Do, Putzel, Martin \emph{et~al.}}]{do}
Do H, Putzel P, Martin AS, \emph{et~al.} (2022).
\newblock \enquote{{Fair Generalized Linear Models with a Convex Penalty}.}
\newblock \emph{Proceedings of Machine Learning Research}, \textbf{162},
  5286--5308.
\newblock 39th International Conference on Machine Learning.

\bibitem[{Dua and Graff(2023)}]{uci}
Dua D, Graff C (2023).
\newblock \emph{{UCI Machine learning repository}}.
\newblock University of California, School of Information and Computer Science.
\newblock \urlprefix\url{https://archive.ics.uci.edu/ml/}.

\bibitem[{{European Commission}(2021)}]{eu2021}
{European Commission} (2021).
\newblock \emph{Proposal for a Regulation Laying Down Harmonised Rules on
  Artificial Intelligence}.
\newblock
  \urlprefix\url{https://digital-strategy.ec.europa.eu/en/library/proposal-regulation-laying-down-harmonised-rules-artificial-intelligence}.

\bibitem[{Fehrman \emph{et~al.}(2017)Fehrman, Muhammad, Mirkes
  \emph{et~al.}}]{fehrman}
Fehrman E, Muhammad AK, Mirkes EM, \emph{et~al.} (2017).
\newblock \enquote{{The Five Factor Model of Personality and Evaluation of Drug
  Consumption Risk}.}
\newblock In \emph{{Data Science}}, pp. 231--242.
\newblock \doi{10.1007/978-3-319-55723-6_18}.

\bibitem[{Friedman \emph{et~al.}(2010)Friedman, Hastie, and
  Tibshirani}]{glmenet}
Friedman J, Hastie T, Tibshirani R (2010).
\newblock \enquote{{Regularization Paths for Generalized Linear Models via
  Coordinate Descent}.}
\newblock \emph{Journal of Statistical Software}, \textbf{33}(1), 1--22.
\newblock \doi{10.18637/jss.v033.i01}.

\bibitem[{Fu \emph{et~al.}(2020)Fu, Narasimhan, and Boyd}]{cvxr}
Fu A, Narasimhan B, Boyd S (2020).
\newblock \enquote{{\pkg{CVXR}: An R Package for Disciplined Convex
  Optimization}.}
\newblock \emph{Journal of Statistical Software}, \textbf{94}(14), 1--34.
\newblock \doi{10.18637/jss.v094.i14}.

\bibitem[{Fukuchi \emph{et~al.}(2013)Fukuchi, Sakuma, and Kamishima}]{fukuchi}
Fukuchi K, Sakuma J, Kamishima T (2013).
\newblock \enquote{{Prediction with Model-Based Neutrality}.}
\newblock In \emph{{Proceedings of the Joint European Conference on Machine
  Learning and Knowledge Discovery in Databases (ECML PKDD)}}, pp. 499--514.
  Springer.
\newblock \doi{10.1587/transinf.2014EDP7367}.

\bibitem[{Fuster \emph{et~al.}(2022)Fuster, {Goldsmith-Pinkham}, Ramadorai, and
  Walther}]{credit}
Fuster A, {Goldsmith-Pinkham} P, Ramadorai T, Walther A (2022).
\newblock \enquote{{Predictably Unequal? The Effects of Machine Learning on
  Credit Markets}.}
\newblock \emph{The Journal of Finance}, \textbf{77}(1), 5--47.
\newblock \doi{10.1111/jofi.13090}.

\bibitem[{Hardt \emph{et~al.}(2016)Hardt, Priceric, and Srebro}]{moritz}
Hardt M, Priceric E, Srebro N (2016).
\newblock \enquote{{Equality of Opportunity in Supervised Learning}.}
\newblock In \emph{Advances in Neural Information Processing Systems},
  volume~29, pp. 3315--3323.

\bibitem[{Hastie \emph{et~al.}(2009)Hastie, Tibshirani, and
  Friedman}]{elemstatlearn}
Hastie T, Tibshirani R, Friedman J (2009).
\newblock \emph{{The Elements of Statistical Learning: Data Mining, Inference,
  and Prediction}}.
\newblock 2nd edition. Springer.
\newblock \doi{10.1007/978-0-387-84858-7}.

\bibitem[{Hort and Sarro(2022)}]{hort}
Hort M, Sarro F (2022).
\newblock \enquote{{Privileged and Unprivileged Groups: An Empirical Study on
  the Impact of the Age Attribute on Fairness.}}
\newblock In \emph{{Proceedings of the International Workshop on Equitable Data
  and Technology}}, pp. 17--24.
\newblock \doi{10.1145/3524491.3527308}.

\bibitem[{Komiyama \emph{et~al.}(2018)Komiyama, Takeda, Honda, and
  Shimao}]{komiyama}
Komiyama J, Takeda A, Honda J, Shimao H (2018).
\newblock \enquote{{Nonconvex Optimization for Regression with Fairness
  Constraints}.}
\newblock \emph{{Proceedings of Machine Learning Research}}, \textbf{80},
  2737--2746.
\newblock {35th International Conference on Machine Learning}.

\bibitem[{Kozodoi and Varga(2021)}]{fairness}
Kozodoi N, Varga TV (2021).
\newblock \emph{{\pkg{fairness:} Algorithmic Fairness Metrics}}.
\newblock \urlprefix\url{https://cran.r-project.org/package=fairness}.

\bibitem[{Lambrecht and Tucker(2019)}]{stem-ads}
Lambrecht A, Tucker C (2019).
\newblock \enquote{{Algorithmic Bias? An Empirical Study of Apparent
  Gender-Based Discrimination in the Display of STEM Career Ads.}}
\newblock \emph{Management Science}, \textbf{65}(7), 2966--2981.
\newblock \doi{10.1287/mnsc.2018.3093}.

\bibitem[{Lang \emph{et~al.}(2019)Lang, Binder, Richter \emph{et~al.}}]{mlr3}
Lang M, Binder M, Richter J, \emph{et~al.} (2019).
\newblock \enquote{{mlr3: A Modern Object-Oriented Machine Learning Framework
  in R}.}
\newblock \emph{Journal of Open Source Software}, \textbf{4}(44), 1903.
\newblock \doi{10.21105/joss.01903}.

\bibitem[{Lohr(2021)}]{lohr}
Lohr SL (2021).
\newblock \emph{{Sampling: Design and Analysis}}.
\newblock 3rd edition. CRC Press.
\newblock \doi{10.1201/9780429298899}.

\bibitem[{Mehrabi \emph{et~al.}(2021)Mehrabi, Morstatter, Saxena
  \emph{et~al.}}]{mehrabi}
Mehrabi N, Morstatter F, Saxena N, \emph{et~al.} (2021).
\newblock \enquote{{A Survey on Bias and Fairness in Machine Learning}.}
\newblock \emph{ACM Computing Surveys}, \textbf{54}(6), 115.
\newblock \doi{10.1145/3457607}.

\bibitem[{Narasimhan(2018)}]{narasimhan}
Narasimhan H (2018).
\newblock \enquote{{Learning with Complex Loss Functions and Constraints}.}
\newblock \emph{Proceedings of Machine Learning Research}, \textbf{84},
  1646--1654.
\newblock 21st International Conference on Artificial Intelligence and
  Statistics.

\bibitem[{{P{\'e}rez-Suay} \emph{et~al.}(2017){P{\'e}rez-Suay}, Laparra,
  {Mateo-Garc{\'i}a} \emph{et~al.}}]{suay}
{P{\'e}rez-Suay} A, Laparra V, {Mateo-Garc{\'i}a} G, \emph{et~al.} (2017).
\newblock \enquote{{Fair Kernel Learning}.}
\newblock In \emph{{Proceedings of the Joint European Conference on Machine
  Learning and Knowledge Discovery in Databases (ECML PKDD)}}, pp. 339--355.
  Springer.
\newblock \doi{10.1007/978-3-319-71249-9_21}.

\bibitem[{Pessach and Shmueli(2022)}]{pessach}
Pessach D, Shmueli E (2022).
\newblock \enquote{{A Review on Fairness in Machine Learning}.}
\newblock \emph{ACM Computing Surveys}, \textbf{55}(3), 51.
\newblock \doi{10.1145/3494672}.

\bibitem[{Pfaff(2022)}]{cccp}
Pfaff B (2022).
\newblock \emph{{\pkg{cccp}: Cone Constrained Convex Problems}}.
\newblock \urlprefix\url{https://cran.r-project.org/package=cccp}.

\bibitem[{Pfisterer \emph{et~al.}(2022)Pfisterer, Siyi, and
  Lang}]{mlr3fairness}
Pfisterer F, Siyi W, Lang M (2022).
\newblock \emph{{\pkg{mlr3fairness:} Fairness Auditing and Debiasing for
  \pkg{mlr3}}}.
\newblock \url{https://mlr3fairness.mlr-org.com},
  \url{https://github.com/mlr-org/mlr3fairness}.

\bibitem[{{R Core Team}(2022)}]{rcore}
{R Core Team} (2022).
\newblock \emph{{R: A Language and Environment for Statistical Computing}}.
\newblock R Foundation for Statistical Computing, Vienna, Austria.
\newblock \urlprefix\url{https://www.R-project.org/}.

\bibitem[{Raghavan \emph{et~al.}(2020)Raghavan, Barocas, Kleinberg, and
  Levy}]{jobs}
Raghavan M, Barocas S, Kleinberg J, Levy K (2020).
\newblock \enquote{{Mitigating Bias in Algorithmic Hiring: Evaluating Claims
  and Practices}.}
\newblock In \emph{{Proceedings of the 3rd Conference on Fairness,
  Accountability and Transparency}}, pp. 469--481.
\newblock \doi{10.1145/3351095.3372828}.

\bibitem[{Schafer \emph{et~al.}(2021)Schafer, {Opgen-Rhein}, Zuber
  \emph{et~al.}}]{corpcor}
Schafer J, {Opgen-Rhein} R, Zuber V, \emph{et~al.} (2021).
\newblock \emph{{\pkg{corpcor:} Efficient Estimation of Covariance and
  (Partial) Correlation}}.
\newblock \urlprefix\url{https://cran.r-project.org/package=corpcor}.

\bibitem[{Scutari(2022)}]{fairml}
Scutari M (2022).
\newblock \emph{{\pkg{fairml}: Fair Models in Machine Learning}}.
\newblock \urlprefix\url{https://cran.r-project.org/package=fairml}.

\bibitem[{Scutari \emph{et~al.}(2022)Scutari, Panero, and Proissl}]{stco21}
Scutari M, Panero F, Proissl M (2022).
\newblock \enquote{{Achieving Fairness with a Simple Ridge Penalty}.}
\newblock \emph{Statistics and Computing}, \textbf{32}, 77.
\newblock \doi{10.1007/s11222-022-10143-w}.

\bibitem[{Simon \emph{et~al.}(2011)Simon, Friedman, Hastie, and
  Tibshirani}]{coxenet}
Simon N, Friedman J, Hastie T, Tibshirani R (2011).
\newblock \enquote{{Regularization Paths for Cox's Proportional Hazards Model
  via Coordinate Descent}.}
\newblock \emph{Journal of Statistical Software}, \textbf{39}(5), 1--13.
\newblock \doi{10.18637/jss.v039.i05}.

\bibitem[{Tay \emph{et~al.}(2023)Tay, Narasimhan, and Hastie}]{tayenet}
Tay JK, Narasimhan B, Hastie T (2023).
\newblock \enquote{{Elastic Net Regularization Paths for All Generalized Linear
  Models}.}
\newblock \emph{Journal of Statistical Software}, \textbf{106}(1), 1--31.
\newblock \doi{10.18637/jss.v106.i01}.

\bibitem[{Tolan \emph{et~al.}(2019)Tolan, Miron, G{\'o}mez, and
  Castillo}]{castillo}
Tolan S, Miron M, G{\'o}mez E, Castillo C (2019).
\newblock \enquote{{Why Machine Learning May Lead to Unfairness}.}
\newblock In \emph{{Proceedings of the Seventeenth International Conference on
  Artificial Intelligence and Law}}, pp. 83--92.
\newblock \doi{10.1145/3322640.3326705}.

\bibitem[{{United Nations}(2015)}]{sdgs}
{United Nations} (2015).
\newblock \emph{{Sustainable Development Goals: The Sustainable Development
  Agenda}}.
\newblock
  \urlprefix\url{https://www.un.org/sustainabledevelopment/development-agenda/}.

\bibitem[{{United Nations}(2023)}]{aiforgood}
{United Nations} (2023).
\newblock \emph{AI for Good}.
\newblock \urlprefix\url{https://aiforgood.itu.int/}.

\bibitem[{{U.S. Bureau of Labor Statistics}(2023)}]{nls}
{US Bureau of Labor Statistics} (2023).
\newblock \emph{{National Longitudinal Surveys of Youth Data Set}}.
\newblock \urlprefix\url{https://www.bls.gov/nls/}.

\bibitem[{Wiśniewski and Biecek(2022)}]{fairmodels}
Wiśniewski J, Biecek P (2022).
\newblock \enquote{{\pkg{fairmodels:} a Flexible Tool for Bias Detection,
  Visualization, and Mitigation in Binary Classification Models}.}
\newblock \emph{The R Journal}, \textbf{14}, 227--243.
\newblock \doi{10.32614/RJ-2022-019}.

\bibitem[{Woodworth \emph{et~al.}(2017)Woodworth, Gunasekar, Ohannessian, and
  Srebro}]{woodworth}
Woodworth B, Gunasekar S, Ohannessian MI, Srebro N (2017).
\newblock \enquote{{Learning Non-Discriminatory Predictors}.}
\newblock \emph{Proceedings of Machine Learning Research}, \textbf{65},
  1920--1953.
\newblock {Conference on Learning Theory}.

\bibitem[{Zafar \emph{et~al.}(2019)Zafar, Valera, Gomez-Rodriguez, and
  Gummadi}]{zafar}
Zafar MB, Valera I, Gomez-Rodriguez M, Gummadi KP (2019).
\newblock \enquote{{Fairness Constraints: a Flexible Approach for Fair
  Classification}.}
\newblock \emph{Journal of Machine Learning Research}, \textbf{20}, 1--42.

\end{thebibliography}

\end{document}